%% file: main.tex
\documentclass[letterpaper, 10 pt, conference]{ieeeconf}  

\IEEEoverridecommandlockouts                              

\usepackage{times}
\usepackage{epsfig}
\usepackage{graphicx}
\usepackage{amsmath}
\usepackage{amssymb}

\usepackage{booktabs}
\usepackage{multirow}
\usepackage{makecell}
\usepackage{bbding}
\usepackage{xcolor}
\usepackage{float}
\usepackage{caption}

\usepackage[ruled]{algorithm2e}

\usepackage{hyperref}

\title{\LARGE \bf
LAGOON: Language-Guided Motion Control
}

\author{Shusheng Xu$^{1,2}{}^*$\thanks{*Equal contribution.}, Huaijie Wang$^{1,2}{}^*$, Yutao Ouyang$^{2,3}$, Jiaxuan Gao$^{1,2}$, Zhiyu Mei$^{1,2}$, Chao Yu$^{1}$, and Yi Wu$^{1,2}$
	\thanks{$^1$ Tsinghua University, Beijing, China
    {\tt\small xuss20@mails.tsinghua.edu.cn, jxwuyi@mail.tsinghua.edu.cn}
 }
    \thanks{$^2$Xiamen University, Xiamen, China}
    \thanks{$^3$Shanghai Qi Zhi Institute, Shanghai, China}
}

\newcommand{\todo}[1]{\textcolor{red}{[TODO: #1]}}
\newcommand{\yw}[1]{\textcolor{red}{[YW: #1]}}

\newcommand{\hide}[1]{}
\newcommand{\name}{{LAGOON}}
\newcommand{\fullname}{{\underline{LA}nguage-\underline{G}uided m\underline{O}tion c\underline{ON}trol}}
\DeclareMathOperator*{\argmax}{arg\,max}
\DeclareMathOperator*{\argmin}{arg\,min}

\begin{document}



\twocolumn[{%
\renewcommand\twocolumn[1][]{#1}
\maketitle
\begin{center}
\captionsetup{type=figure}
\includegraphics[width=1\linewidth]{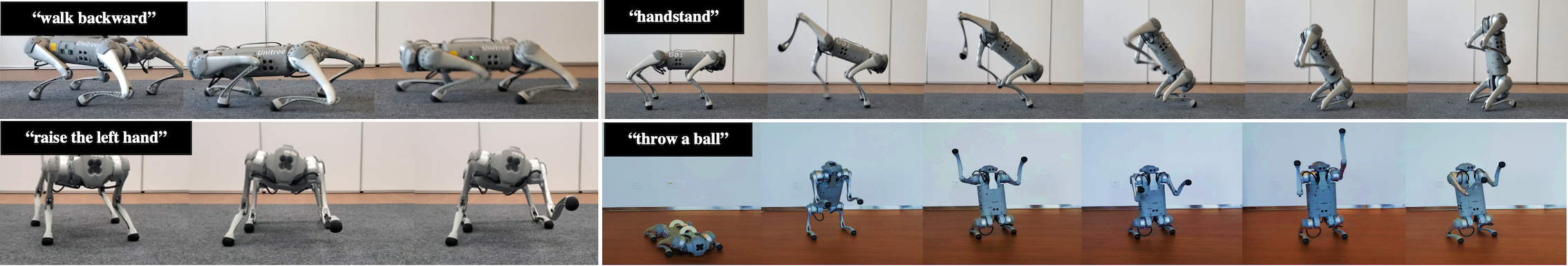}
\captionof{figure}{We developed a system called {\name}. Given a \emph{high-level} language command, {\name} can autonomously train a policy to control the quadrupedal robotic to take actions according to the provided command.}
\label{fig:teaser}
\end{center}
}]

\renewcommand{\thefootnote}{\relax}
\footnotetext{*Equal contribution.}
\footnotetext{$^1$ Tsinghua University, Beijing, China
    {\tt\small xssstory@gmail.com}
 }
    \footnotetext{$^2$ Shanghai Qi Zhi Institute, Shanghai, China
    }
    \footnotetext{$^3$ Xiamen University, Xiamen, China}
    
\renewcommand{\thefootnote}{\arabic{footnote}}


\input{00_abs.tex}

\input{10_intro.tex}

\input{20_related.tex}
\input{30_prelim}

\input{40_method.tex}
\input{50_exper.tex}

\input{60_con.tex}





\section*{ACKNOWLEDGMENT}
Yi Wu is supported by 2030 Innovation Megaprojects of China(Programme on New Generation Artificial Intelligence) Grant No. 2021AAA0150000. 




\bibliographystyle{IEEEtran}
\bibliography{IEEEabrv,egbib}

\end{document}


\title{Language-Guided Generation of Physically Realistic Robot Motion and Control \\ Supplementary Material}

\author{First Author\\
Institution1\\
Institution1 address\\
{\tt\small firstauthor@i1.org}
\and
Second Author\\
Institution2\\
First line of institution2 address\\
{\tt\small secondauthor@i2.org}
}

\maketitle
\ificcvfinal\thispagestyle{empty}\fi

\input{appendix}
{\small
\bibliographystyle{ieee_fullname}
\bibliography{egbib}
}

%% file: 00_abs.tex
\begin{abstract}

We aim to control a robot to physically behave in the real world following any high-level language command like ``cartwheel'' or ``kick''. 
Although human motion datasets exist, this task remains particularly challenging since generative models can produce physically unrealistic motions, which will be more severe for robots due to different body structures and physical properties. 
Deploying such a motion to a physical robot can cause even greater difficulties due to the sim2real gap.
We develop {\fullname} (\name), a multi-phase reinforcement learning (RL) method to generate physically realistic robot motions under language commands. {\name} first leverages a pretrained model to generate a human motion from a language command. Then an RL phase trains a control policy in simulation to mimic the generated human motion. Finally, with domain randomization, our learned policy can be deployed to a quadrupedal robot, leading to a quadrupedal robot 
that can take diverse behaviors in the real world under natural language commands.
    
\end{abstract}

%% file: 10_intro.tex
\section{Introduction}

Reinforcement learning (RL) has been a trending paradigm for addressing intricate challenges in robotic control, encompassing domains such as bipedal~\cite{li2023robust} and quadrupedal locomotion~\cite{margolis2022walktheseways}, drone racing~\cite{kaufmann2023champion}, and robotic arm manipulation~\cite{li2021learning}. Specifically, an RL-based approach trains a neural policy within a simulated environment through the formulation of task-specific reward functions~\cite{ibarz2021train}, followed by the transfer of the acquired policy to a physical robot via domain randomization techniques~\cite{peng2018sim}.


Despite the successes, most existing RL methods focus on low-level robust control tasks, such as walking~\cite{rudin2022learning}, and rely on a heavily engineered task-specific reward~\cite{margolis2022walktheseways,rudin2022learning}.
Whenever the task goal changes, the creation of a new reward function requires substantial efforts.
It remains an open challenge whether we can directly train control policies to generate complex behaviors according to \emph{high-level} \emph{semantic} commands, such as ``throw a ball'' or ``handstand'', without the need to specify any sophisticated reward function.


The prospect of controlling robots through high-level language commands becomes increasingly promising with the advancement of pre-trained models~\cite{saycan2022arxiv}. 
Recently, with diffusion models, it has become possible to generate diverse human trajectories based on high-level language commands~\cite{tevet2022human}, which suggests a feasible direction for language-guided control: initiating motion generation using a pre-trained model followed by the implementation of imitation learning.
However, a common pitfall of the existing motion generation methods is that the generated motion may often violate real-world physical constraints since no physics simulation is performed during such an end-to-end generation process. 
When generating robot motions, this issue is more severe,
since most existing motion datasets are collected from human demonstrators while a robot can have a drastically different body structure from humans. Further applying such a motion to a physical robot can introduce even greater challenges due to the sim2real gap.

We propose a novel RL-based method, {\fullname} ({\name}), to address all the aforementioned challenges. {\name} is a multi-stage approach benefiting from both motion generation and RL training. First, {\name} adopts a motion diffusion model to generate a human motion from a language command. 
Then, we convert the generated human motion into a semantically desired yet physically unrealistic robot motion. Next, an RL phase is performed to learn a policy in a physics engine to control a simulated robot to mimic the target motion. 
Finally, with domain randomization, the learned RL policy can be deployed on a real-world robot to physically perform the motion.

We emphasize that effectively training an RL policy to mimic a target motion is non-trivial in our setting. 
Existing algorithms that can learn motion control from demonstration videos often assume \emph{perfect} demonstrations produced by human professionals~\cite{pmlr-v155-schmeckpeper21a,sermanet2018time,zakka2021xirl,pmlr-v162-seo22a}.
In contrast, our target motion is completely synthetic and can be highly unrealistic, including physically impossible poses or even missing frames that lead to teleportation or floating behaviors.
We adopt a special rewarding mechanism for motion imitation, which combines both adversarially learned critic reward for high-level semantic consistency and an optimal-matching-based state-error reward to enforce fine-grained consistency to key frames from the target motion. 


We evaluate {\name} on two types of robots in simulation: a humanoid robot, which resembles the human body structure for easier motion imitation, and a quadrupedal robot, posing more challenges for RL training due to its distinct body structure. Our empirical findings demonstrate that {\name} is able to generate robust control policies for both robots, producing physically realistic behaviors following various language commands.
For the humanoid robot, {\name} produces better policies on high-level commands like ``cartwheel'' as well as commands requiring fine-grained control like ``kick'', outperforming all the RL baselines consistently. 
For the quadrupedal robot, despite the substantial differences in body structure, {\name} is able to produce a policy to execute challenging commands like ``\emph{throw a ball}'' by controlling the robot to stand up and wave its front legs, just like a human.

We also successfully deployed our system to a physical quadrupedal robot in the real world, resulting in a quadrupedal robot automatically performing diverse motions that are semantically consistent with a variety of language commands, such as ``walk backward'', ``handstand'', ``raise the left hand'' and ``throw a ball'', as shown in Fig.~\ref{fig:teaser}.

%% file: 20_related.tex
\section{Related work}

\subsection{Language-Conditioned Motion Generation}
Early attempts at translating text descriptions into human motion employed deterministic encoder-decoder architectures~\cite{ahuja2019language2pose,ghosh2021synthesis}.
Recent efforts have shifted towards deep generative models such as GANs, VAEs~\cite{guo2022generating,petrovich2022temos}, or diffusion models~\cite{ren2022diffusion,tevet2022human,zhang2022motiondiffuse} due to the stochastic nature of motions. Note that these motion generation methods are trained on large datasets typically limited to human motions.
Despite their superior performances, standard deep generative models do not explicitly incorporate the law of physics into the generation process. 
\cite{yuan2022physdiff} integrate the imitation policy trained in a physics simulator into the sampling process of the diffusion model. However, this approach still relies on a manually designed residual force~\cite{yuan2020residual} at the root joint to compensate for the dynamics mismatches between the physics model and real humans, which is not suitable for real robot control. In comparison, we concentrate on physically realistic robots with various structures.

\subsection{Learning Methods for Robot Control}
To attain natural behaviors, researchers carefully designed heuristics for symmetry~\cite{yu2018learning}, energy consumption~\cite{Al2013trajectory}, and proper contact with the environment~\cite{mordatch2012discovery}. Nonetheless, these methods typically demand substantial domain expertise and are thus limited to simpler tasks. In contrast, imitation learning (IL) is more general. It can learn from expert demonstrations and deploy the learned policies in physics simulators~\cite{peng2018deepmimic,bergamin2019drecon,peng2021amp} or the real world ~\cite{peng2020learning,peng2022adversarial}.
IL assumes \emph{perfect} real-world data, while we only have access to \emph{imperfect} synthetic demonstrations.
Some approaches use pre-trained models for robot control, such as using a language model for representation learning~\cite{shridhar2022cliport} or semantic planning~\cite{ahn2022can}. However, these methods demand human demonstrations for low-level control, while we do not require additional control data.
Others~\cite{ajay2022conditional,dai2023learning} train a video diffusion model to generate a sequence of trajectory states and use inverse dynamics to infer actions. In our setting, the reference motion and the actual policy trajectory cannot be precisely aligned, making it infeasible to produce actions through inverse dynamics.

\subsection{State-Based Imitation Learning}
In cases where expert actions are unavailable, policies have to learn from states. One approach is to train a dynamics model to predict actions from state transitions and then apply behavior cloning~\cite{BCO, pmlr-v97-edwards19a}. Other methods perform RL directly with a state-based imitation reward, such as differences in state representations~\cite{bergamin2019drecon,peng2018deepmimic,sermanet2018time,zakka2021xirl} or an adversarially learned discriminator ~\cite{merel2017learning,torabi2018generative,karnan2022adversarial,peng2021amp,Liu2020State,Gangwani2020State-only,viano2022robust}.
We leverage both of these reward types to train our RL policy. Since the policy motion and the reference motion can be largely mismatched in our setting, both reward terms are critical for the empirical success.

\subsection{Motion and Control of Quadrupedal Robots}
Previous approaches have focused on generating controllable or natural quadrupedal motions and gaits either by imitating animals~\cite{peng2020learning} or by relying on heavily engineered reward functions~\cite{margolis2022walktheseways,rudin2022learning}. \cite{kim2022human} build a human-to-quadrupedal control interface by collecting matching pairs of human and robot motions. Our work demonstrates the ability to generate diverse motions without the need for domain-specific data or meticulously designed reward functions.

%% file: 30_prelim.tex
\section{Preliminary}

We consider a robot control problem following language commands. The aim is to derive a control policy $\pi$ conditioned on language command $c$.

\subsection{Human Motion Generation}
Human motion generation aims to produce a sequence $x_{0:H} = \{x_h\}_{h=0}^H$. $x_h \in \mathbb{R}^{J\times K}$ represents a human pose using $K$-dimensional features of $J$ joints. Here $K$-dimensional features can be either the joint angles or positions. 
A language-conditioned motion generation model aims to generate a motion matching the language command $c$.

Diffusion models~\cite{ho2020denoising,sohl2015deep,song2020denoising} are able to generate high-quality human motions~\cite{ren2022diffusion,tevet2022human,zhang2022motiondiffuse}.
These works model the data distribution by injecting noises into it and gradually denoise a sample from a Gaussian distribution. The forward diffusion process injects i.i.d. Gaussian noises, namely 
\begin{equation*}
    x_{0:H}^l\sim\mathcal N(\sqrt{\alpha_l}x_{0:H}^{l-1},(1-\alpha_l)\mathbf I),
\end{equation*}
where $x_{0:H}^0$ denotes samples drawn from the real data distribution $p^0(x^0)$. For large enough $l$, $x_{0:H}^l$ approximately follows the Gaussian distribution $\mathcal N(\mathbf 0,\mathbf I)$. A denoiser $F_d(x^l,l)$ is trained to gradually denoise $x_{0:H}^l$ back to $x_{0:H}^{l-1}$. The training objective of $F_d$ is usually given by
\begin{equation*}
    \mathbb{E}_{x^0\sim p^0(x^0), l\sim q(l), \epsilon\sim \mathcal{N}(\mathbf{0}, \mathbf{I})}
    \left [ 
        \lambda(l)\|x^0 - F_d(x^l, l)\|^2
    \right],
\end{equation*}
where $q(l)$ is a distribution from which $l$ is samapled and $\lambda(l)$ is a weighting factor.

\subsection{RL for Control}
Rather than generating a motion directly, RL methods learn a policy in simulation to control a robot to perform the desired motion according to some given reward function.

\subsubsection{Markov Decision Process}

The robot control problem can be formulated as a Markov Decision Process (MDP) denoted by $M=\langle\mathcal S,\mathcal A, \mathcal T, r,\gamma\rangle$. Here $\mathcal S$ is the state space, and $\mathcal A$ is the action space. $\mathcal T:\mathcal S\times\mathcal A\times\mathcal S\to[0,1]$ 
is the transition function. $\mathcal T(s'|s,a)$ denotes the probability of reaching state $s'$ from state $s$ under action $a$. $r :\mathcal S\times \mathcal A\to\mathbb R$ is the reward function and $\gamma$ is the discounted factor. In our task setup, the reward function $r$ is generated based on a language command $c$. In practice, when applying RL for robot control, complex reward designs are often required due to a lot of motor joints and movement constraints~\cite{rudin2022learning}.



At each time step $t$, the control policy produces an action $a_t \sim \pi(\cdot|s_t)$, and receives a reward $r(s_t,a_t)$. The objective of RL is to find the optimal policy $\pi^\star$ that could maximize the discounted accumulated reward,
\begin{equation*}
    \label{eq.rl}
    \pi^\star=\arg\max_{\pi} \mathbb{E}\left[\sum_{t \geq 0} \gamma^t r(s_t, a_t) \mid a_t \sim \pi\left(\cdot \mid s_t\right), s_0\right]
\end{equation*}
where $s_0$ is the initial state.

\subsubsection{RL-Based Motion Imitation}


Our goal is to derive an RL policy given a language command $c$. However, it is difficult to design reward functions directly for robot control tasks. One approach to solving this problem is imitation learning (IL), which is also called learning from demonstration. In imitation learning, we assume that there is a dataset $\mathcal{D} = \{(s_i, a_i) \}_{i=1}^{m}$ collected by an expert or reference policy $\pi_E$. 
The goal of IL is to find the optimal policy $\pi^*$ that covers the distribution of state-action pairs in the dataset $\mathcal{D}$.
Two typical approaches are behavior cloning~\cite{ho2016generative} and inverse reinforcement learning~\cite{bojarski2016end}, which match state-action pairs between the expert and the imitator. 
We are interested in cases where expert actions are not available and the dataset only contains a trajectory of states, i.e. $\mathcal{D}=\{x_i\}_{i=0}^H$.
One approach to solve this problem is state-based IL~\cite{merel2017learning,peng2018deepmimic,peng2021amp}. State-based IL typically designs a state-error reward~\cite{peng2018deepmimic}, which encourages the imitator to reach the reference states $x_{0:H}$. 
Let $s_t$ denote the imitator state at timestep $t$, the state-error reward $r^\mathrm{err}_{t}$ is defined as,
\begin{equation}
\label{eq.err}
    r^\mathrm{err}_t = Sim(x_t, s_t)
\end{equation}
where $Sim(x_t, s_t)$ represents the similarity between the reference state $x_t$ and the imitator state $s_t$ at timestep $t$.

A crucial assumption of state-based IL is a strict timing alignment between the demonstration and the rollout trajectory, i.e. the agent receives a high reward at timestep $t$ if and only if $s_t$ is close to $x_t$. This  can be problematic when the reference motion $x_{0:H}$ is not physically realistic. Adversarial imitation learning (AIL) tackles this issue by training a discriminator $D_\psi$ to differentiate behaviors generated by the imitator from the reference motion $x_{0:H}$, where $\psi$ is the network parameters. 
$D_\psi$ then scores the states $s_t$ generated by the imitator. 
Specifically, $D_\psi$ is trained to discriminate between state transitions in the reference motion and the samples generated by the current policy $\pi_\theta$:
\begin{multline*}
    \mathcal L_\text{disc}=\mathbb E_{h}[-\log D_\psi(x_h, x_{h+1})] \\
    +\mathbb E_t [-(1-\log D_\psi(s_t,s_{t+1}))]+w_\text{gp}\mathcal L_\text{gp},
\end{multline*}
where $\mathcal L_\text{gp}$ is a gradient penalty term given by
\begin{small}
\begin{equation*}
    \mathcal L_\text{gp}=\mathbb E_h[\|\nabla_{(x_h,x_{h+1})}D_\psi(x_h, x_{h+1})\|^2].
\end{equation*}
\end{small}
As discussed in \cite{Mescheder2018WhichTM}, this zero-centered gradient penalty stabilizes training and helps convergence. AIL optimizes the policy $\pi_\theta$ to maximize the discounted accumulated adversarial reward. The adversarial reward is given by,
\begin{equation}
\label{eq.adv_rew}
    r^\mathrm{adv}_t = -\log(1-D_\psi(s_{t-1},s_t)).
\end{equation}

%% file: 40_method.tex
\section{Methodology}

\begin{figure}
    \centering
    \vspace{1mm}
    \includegraphics[width=1\linewidth]{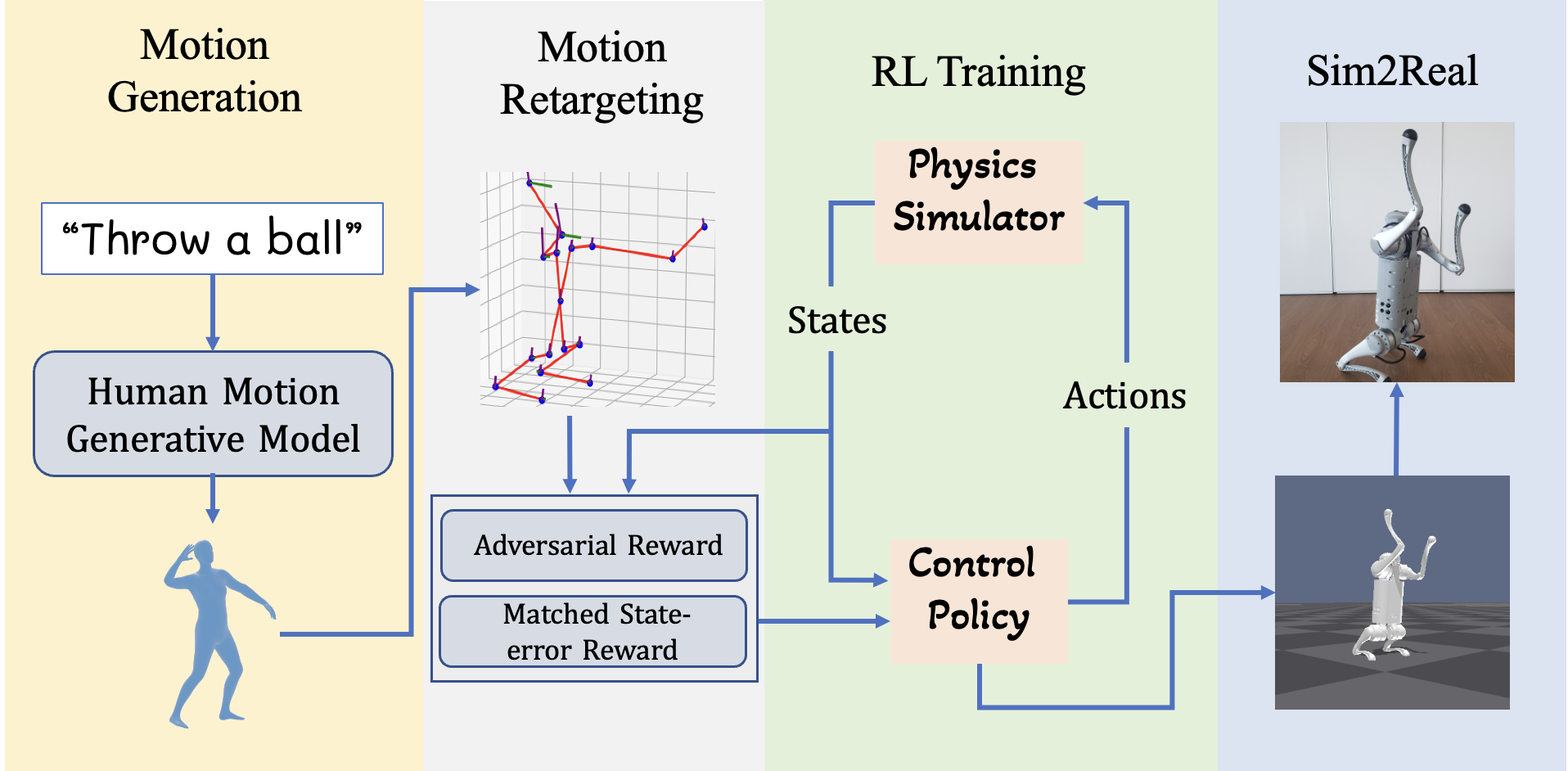}
    \caption{Overview of the multi-phase method {\fullname} ({\name}). To make a robot follow a language command such as ``throw a ball'', we first generate a human motion using the motion generation model. Then the human motion can be retargeted to a robot skeleton that differs largely from humans. By introducing RL training, we train a robust control policy in the physics simulator. Finally, we deploy the control policy to the real-world robot.}
    \label{fig:pipeline}
\vspace{-6mm}
\end{figure}

As shown in Fig.~\ref{fig:pipeline}, we derive a robot control policy following a language command $c$ through a multi-phase method. We first generate a motion sequence $x_{0:H}$ conditioned on $c$ using a human motion generation model. $x_{0:H}$ is then retargeted to the robot skeleton to produce a robot motion $y_{0:H}$. Finally, we adopt RL training to obtain a control policy $\pi_\theta$ and transfer the learned policy to the real world via domain randomization.

\subsection{Motion Generation and Motion Retargeting}
\label{sec.lcmg}
\begin{figure}[ht]
    \centering
    \includegraphics[width=.9\linewidth]{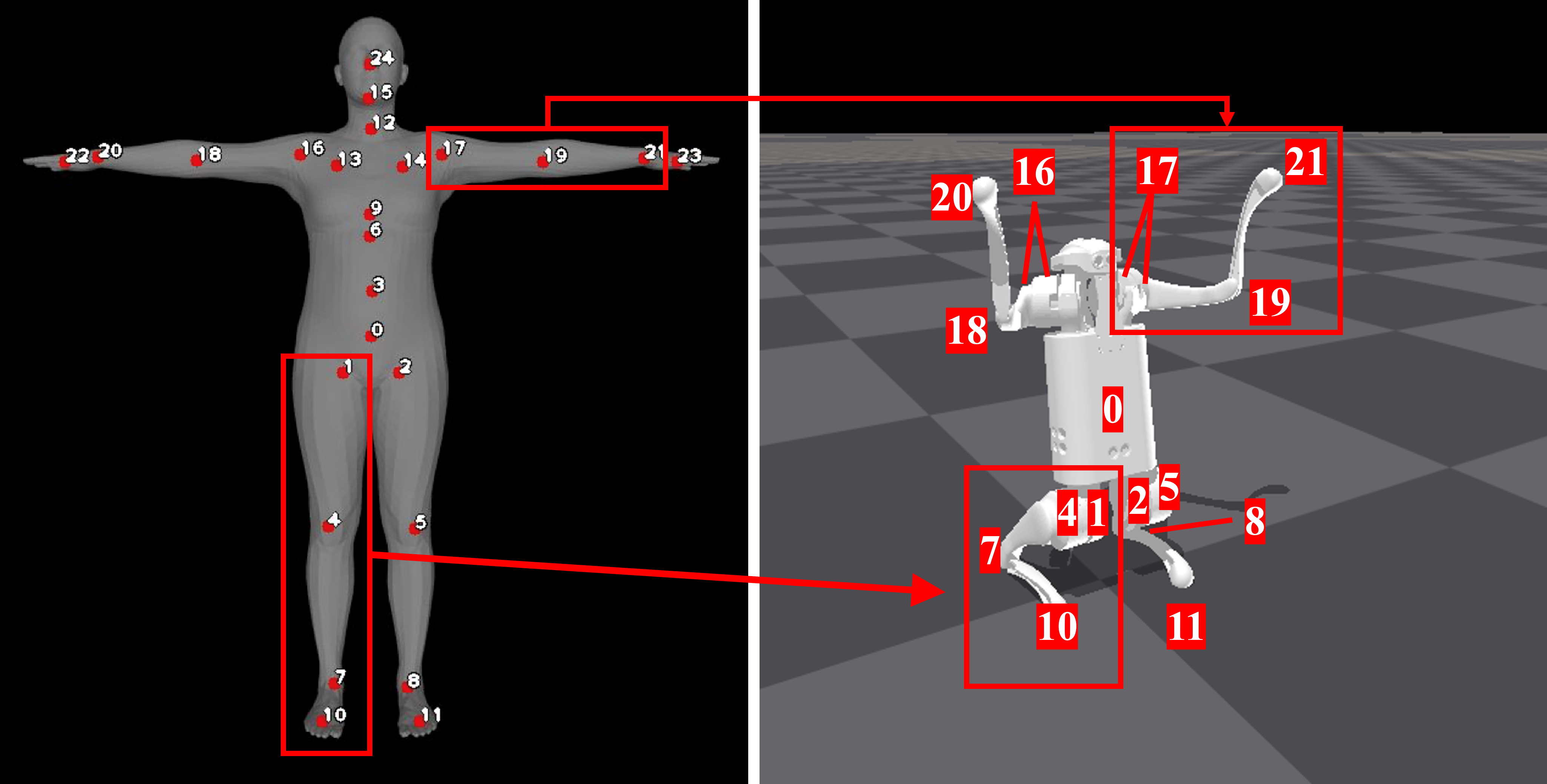}
    \caption{Joint mappings from the SMPL skeleton to the quadrupedal robot. Note that 16 and 17 in the quadrupedal skeleton each correspond to 2 joints.}
    \label{fig:joint_mapping}
    \vspace{-6mm}
\end{figure}

In the motion generation stage, we adopt a SOTA Human Motion Diffusion Model (MDM)~\cite{tevet2022human} to generate human motion $x_{0:H}$ conditioned on a language command $c$. Since MDM can only generate human motion, we then adopt a retargeting stage to map the human motion $x_{0:H}$ to the desired robot motion $y_{0:H}$. Taking the quadrupedal robot as an example, we map the human skeleton to the robot's skeleton, with the human arms corresponding to the robot's two front legs and the human legs corresponding to the robot's two rear legs, and then retarget each joint and joint rotation (see Fig~\ref{fig:joint_mapping}). For the humanoid robot, the number of joints on different skeletons may vary and therefore also need to be retargeted accordingly. For those joints that are redundant, they are simply discarded\footnote{\url{https://github.com/NVIDIA-Omniverse/IsaacGymEnvs/tree/main/isaacgymenvs/tasks/amp/poselib}}.


\subsection{RL Training}
\label{sec.rl}
The RL phase trains a control policy $\pi_\theta$ to imitate the retargeted robot motion $y_{0:H}$. The policy takes in robot states and outputs an action to interact with the physics simulator. The reward is calculated by comparing the robot states with the retargeted reference robot motion. 
The motion generated by MDM  is \emph{physics-ignoring}, so floating, penetration and teleportation behaviors  may often occur, which can be amplified in the retargeted robot motion $y_{0:H}$. Inaccurate reference motions pose significant challenges to motion imitation, which we tackle via a careful reward design.


\subsubsection{Reward Design}

We combine both the adversarial reward $r^\mathrm{adv}$ (Eq.~(\ref{eq.adv_rew})) and a variant of the state-error reward $r^\mathrm{me}$ (Eq.~(\ref{eq.me_rew})). Intuitively, the adversarial reward is universal to capture high-level semantic consistency with the reference motion. However, we empirically observe that only using an adversarial reward can fail to match critical poses in the reference motion. For example, when given the command ``kick'', the policy trained with the adversarial reward alone only learns to stand but fails to perform a kick. 
Hence, for more fine-grained body control, we additionally leverage a state-error reward, which can be nontrivial since 
 the policy trajectory and the reference motion are not well aligned. 

To tackle the temporal mismatch issue, we employ a matching algorithm between the policy rollout trajectories and the reference motion to find the best temporal alignment leading to the highest state-error reward. 
More specifically, let $y_{0:H}$ be the reference motion sequence and $\tau=(s_0,s_1,\ldots,s_T)$ be a trajectory from the policy. We define a matching $M$ between $y_{0:H}$ and $\tau=(s_0,s_1,\ldots,s_T)$ by
\begin{multline*}
    M=\{(u_{0:k},v_{0:k})\mid k\in\mathbb N_+, \\
    0\le u_0 < \cdots < u_k\le H, 0\le v_0 < \cdots < v_k\le T\}.
\end{multline*}
where $s_{v_m}$ is matched with $y_{u_m}$ for all $0\le m \le k$. Recall that $Sim(y_i,s_j)$ is a similarity measurement between the $i$-th motion state $y_i$ and the robot state $s_j$ at timestep $j$. 
We aim to find the optimal matching $M^*=(u^*_{0:k^*},v^*_{0:k^*})$ that maximizes the total similarity, namely
\begin{equation*}
    (u^*_{0:k^*}, v^*_{0:k^*})=\argmax_{(u_{0:k},v_{0:k})\in M}\sum_{m=0}^k Sim(y_{u_m},s_{v_m}), 
\end{equation*}
which can be solved via dynamic programming. The optimal matching $M^*$ helps filter out unrealistic poses and allows the robot to smoothly transit between two consecutive motion frames. 
With the optimal matching $(u^*_{0:k^*},v^*_{0:k^*})$, the matched state-error reward is defined as
\begin{equation}
    \label{eq.me_rew}
    r^\mathrm{me}_t = Sim(y_{u^*_m}, s_{v^*_m}) \cdot \mathbb{I}(t = v^*_m \in v^*_{0:k^*})
\end{equation}

Our final reward is a combination of adversarial reward and matched state-error reward. 
\begin{equation}
    \label{eq.final_rew}
    r_t = \lambda_\mathrm{adv} r^\mathrm{adv}_t + \lambda_\mathrm{me} r^\mathrm{me}_t
\end{equation}

where $\lambda_\mathrm{adv}$ and $\lambda_\mathrm{me}$ are weighting factors. We also remark that the adversarial reward remains critical since it provides much denser reward signals than the state-error reward.

\subsubsection{PPO with Augmented Critic Inputs}

We utilize PPO~\cite{PPO} for RL training, which adopts an actor-critic structure with two separate neural networks, i.e. a policy $\pi_{\theta}$ and a value function $V_{\phi}$. The critic is only used for variance reduction at training time, so we can input additional information not presented in robot states to the value network to accelerate training. 
In particular, given the trajectory $\tau$, the reference sequence $y$, and the optimal matching $M^*$, for each state $s_t$, we take the next future reference motion for $s_t$ from $M^*$ as the additional information to the critic. Such future information significantly improves training in practice. 
We also remark that similar techniques have been widely adopted in  multi-agent RL~\cite{yu2021surprising}.
\subsubsection{Robust control with domain randomization}
In order to learn robust control policies, we adopt domain randomization~\cite{peng2018sim} during RL training.
We randomize both the terrains and physics parameters in the simulator so that the trained RL policy can generalize to different terrain conditions and even to the real world.

%% file: 50_exper.tex
\section{Experiment}

We conduct experiments on the humanoid and quadrupedal robots in the IsaacGym~\cite{makoviychuk2021isaac} simulator. We test \name{} using the 28 DoF humanoid from AMP~\cite{peng2021amp,makoviychuk2021isaac} and the go1 quadrupedal robot\footnote{\url{https://www.unitree.com/go1}} with 12 DoF.  We use the target DoF angles of proportional derivative (PD) controllers as the actions. The action dimensions are 28 and 12 for the humanoid and quadrupedal robots, respectively. 
We also deploy the policies in the real-world quadrupedal robot.


\begin{figure}[t]
    \centering
    \centering
    \vspace{1mm}
    \includegraphics[width=.84\linewidth]{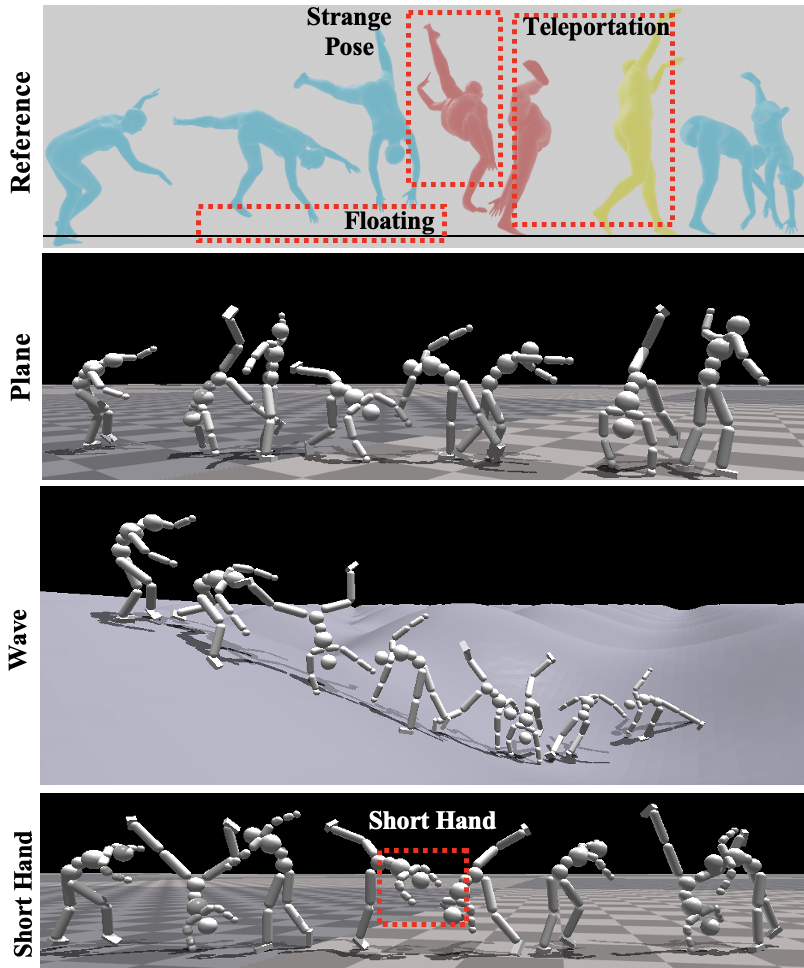}
  \caption{The reference motion sequence overlooks the law of physics. The trained policies robustly perform the ``cartwheel'' motion even in complex terrains (Wave) or a different skeleton with shorter arms (Short Hand).
  }
  \label{fig:human_cartwheel}
    \vspace{-8mm}
\end{figure}


\subsection{Humanoid Robot}

For the humanoid robot, We conduct experiments on the tasks specified by the texts ``the person runs backward'', ``cartwheel'', and ``the person kicks with his left leg''. These tasks include the movement and rotation of the entire body and fine-grained control of parts of the body.

\textbf{Baselines:} We compare \name{} with other state-based imitation learning methods. BCO~\cite{BCO} is a non-RL method that learns an inverse dynamics model to label the reference motion with actions and adopt behavior cloning on the labeled motion. Other baselines are RL-based methods. State Err. only uses the state-error reward.
GAILfO~\cite{torabi2018generative,karnan2022adversarial,peng2021amp} and RGAILfO~\cite{viano2022robust} utilize the adversarial rewards. In particular, RGAILfO tried to alleviate the problem of dynamics mismatch by introducing an adversary policy when collecting trajectories. 
We also conduct experiments using the hand-designed reward and denote it as pure RL. The reward for ``run backward'' is the backward velocity at each timestep. For the task ``kick'', let $h_t$ denote the height of the robot's left foot at timestep $t$, and the reward is computed as $\max(0, h_t - \max_{s<t}(h_s))$, which measure the difference between the current foot height and the previous maximum height. The ``cartwheel'' task is excluded since it is difficult to manually design reward functions for doing cartwheels.

\textbf{Training Details:} The input states of the control policy include the root's linear velocity and angular velocity, the local velocity and rotation of each joint, and the 3D position of the end-effectors. We train the policies by randomizing the terrains and the physics parameters to handle various complex situations. There are four terrains during training. ``Plane'' refers to a flat surface without variations in elevation. ``Rand.'' is terrain with a bit of random undulation. ``Pyramid'' is square cone terrain with steps. ``Wave'' is the terrain with a great deal of undulation. We additionally train a policy for a humanoid robot with shorter arms. 

We create 4096 parallel simulation environments in IsaacGym to collect training samples. The max episode length of each simulation is 300. Each environment would be reset when the robot in it falls (i.e., any part of the humanoid except hands and feet is in contact with the ground). We train the policy for 5,000 iterations and adopt the final policy for evaluation. 

\input{table}

\subsubsection{Illustration of Learned Motion}
A critical problem of the generated motion is physics ignoring. As shown in Fig.~\ref{fig:human_cartwheel}. The top picture is the motion generated by MDM conditioned on the language prompt ``cartwheel'', where some postures are floating or ground-penetrating. We mark the strange postures impossible to imitate in red, and the posture represents that teleportation occurs in yellow. After retargeting and RL training, the control policy can do cartwheels in various terrains. For example, the control policy can do cartwheels on a large slope.
We also train the policy on the robot with short hands. We can observe that the robot can also do cartwheels steadily.
Since these behaviors are conducted in the physics simulator, the generated robot motion would always be physically realistic. 
\subsubsection{Comparision with Baselines} \quad


We compare \name{} with various RL-based baselines.
The results are listed in Tab.~\ref{tab.compare}. We evaluate the policies on each task using success rates. 
The BCO baseline performs worst on all tasks and all terrains, as it is challenging to estimate the environment dynamics. Pure RL policies achieve high success rates. However, we observe the policy can't maintain balance. And it is usually difficult to design the reward for the pure RL method. \name{} consistently outperforms the baselines on different terrains.
For the task ``cartwheel'', the low success rates of State Err. indicate that tracking the states in the reference motion alone may not suffice for completing complex skills. 
For the task ``kick'', methods without state-error rewards (GAILfO, RGAILfO) have significantly lower success rates than methods with state-error rewards (\name{}, State Err.).  This result suggests that the state error reward encourages the policy to imitate the fine-grained poses from the reference.

\subsection{Quadrupedal Robot}
\begin{figure}[ht]
    \centering
    \vspace{-4mm}
    \includegraphics[width=0.9\linewidth]{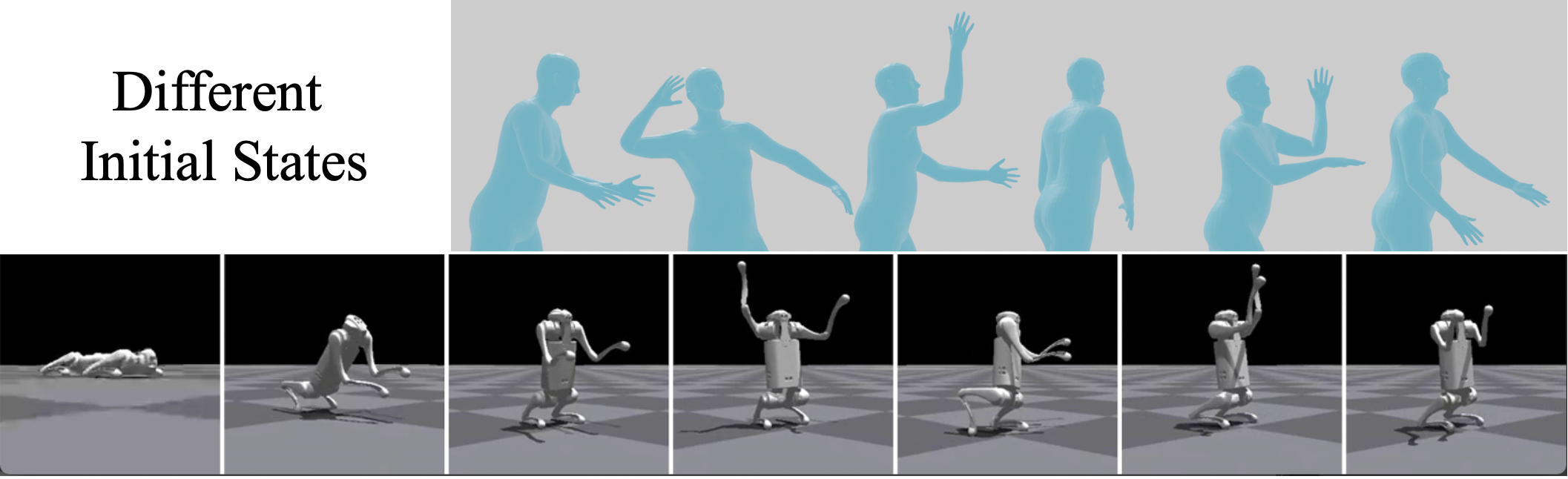}
    \caption{The task of ``throw a ball'' on the quadrupedal robot. Even though the getting up does not appear in the reference motion, the quadrupedal robot learns how to get up from the ground and then wave its front legs.}
    \label{fig:dog}
    \vspace{-4mm}
\end{figure}
We evaluate \name{} on the task conditioned on the language text ``the person throws a ball''. The quadrupedal robot has an essentially different body structure from the humans, and the initial states of the robots and the reference motion are largely different. The robot must first learn to ``stand'' like the humans without extra data. 
The input states of the quadrupedal robot policy include the projected gravity, the local velocity and rotation of each joint, and the actions taken in the last timestep. 
The result in Fig.~\ref{fig:dog} demonstrates that \name{} makes the policy ``stand'' on two feet, then the robot successfully takes the behavior ``throws a ball''. 

\subsection{Real-World Robot Deployment}

We also deploy the trained policy in the real-world quadrupedal robot.
For the real-world experiments, besides the reward mentioned in Eq.~\ref{eq.final_rew}, we add the following auxiliary rewards during the training process to protect the robots:
\begin{equation}
    r_{t}^\mathrm{aux} = [r_{t}^\mathrm{pl}, r_{t}^\mathrm{a},r_{t}^\mathrm{tor}, r_{t}^\mathrm{ar}, r_{t}^\mathrm{col}, r_{t}^\mathrm{slip}]
 \end{equation}

where $r_{t}^\mathrm{pl}$ is a strict penalty to prevent the motor position from exceeding its limit. $r_{t}^\mathrm{a}$ and $r_{t}^\mathrm{tor}$ are used to mitigate motor behavior by penalizing excessive acceleration and torque. $r_{t}^\mathrm{ar}$ is introduced with the purpose of smoothing the action of the control policy. To enhance motion regulation and safety, we incorporate $r_{t}^\mathrm{col}$ to penalize collisions. Additionally, we introduce $r_{t}^\mathrm{slip}$ to penalize the velocity component of the robot's toe perpendicular to the ground normal when it makes contact with the ground. This measure significantly alleviates slip-related issues. Furthermore, we incorporate a contact reset mechanism when the robot falls. We also adopt domain randomization, the details are listed in Table~\ref{tab:dog_rand}.

\begin{table}[ht]
    \centering
    \scriptsize
    \vspace{1mm}
    \caption{Randomization parameters.}
    \vspace{-2mm}
    \begin{tabular}{|c|c|c|c|c|}
        \hline
        Parameter & Operation & Distribution & Unit \\
        \hline
        Obs.Gravity & Additive & $\operatorname{U}(-0.05,0.05)$ & $\mathrm m/\mathrm s^2$ \\
        Obs.ROT & Additive & $\operatorname{U}(-0.01, 0.01)$ & $\mathrm{rad}$\\
        Obs.VEL & Additive & $\operatorname{U}(-1.5, 1.5)$ & $\mathrm{rad}/\mathrm s$ \\
        Trunk Mass & Additive & $\operatorname{U}(-1, 1)$ & $\mathrm{kg}$ \\
        Body Friction & Scaling
        & $\operatorname{U}(0.3, 3)$ & 1 \\
        Proportional Gain & Scaling & $\operatorname{U}(0.7, 1.3)$ & 1 \\
        Derivative Gain & Scaling & $\operatorname{U}(0.7,1.3)$ & 1 \\
        \hline
    \end{tabular}
    \label{tab:dog_rand}
    \vspace{-6mm}
\end{table}

\begin{figure}[ht]
    \centering
    \vspace{-3mm}
    \includegraphics[width=1\linewidth]{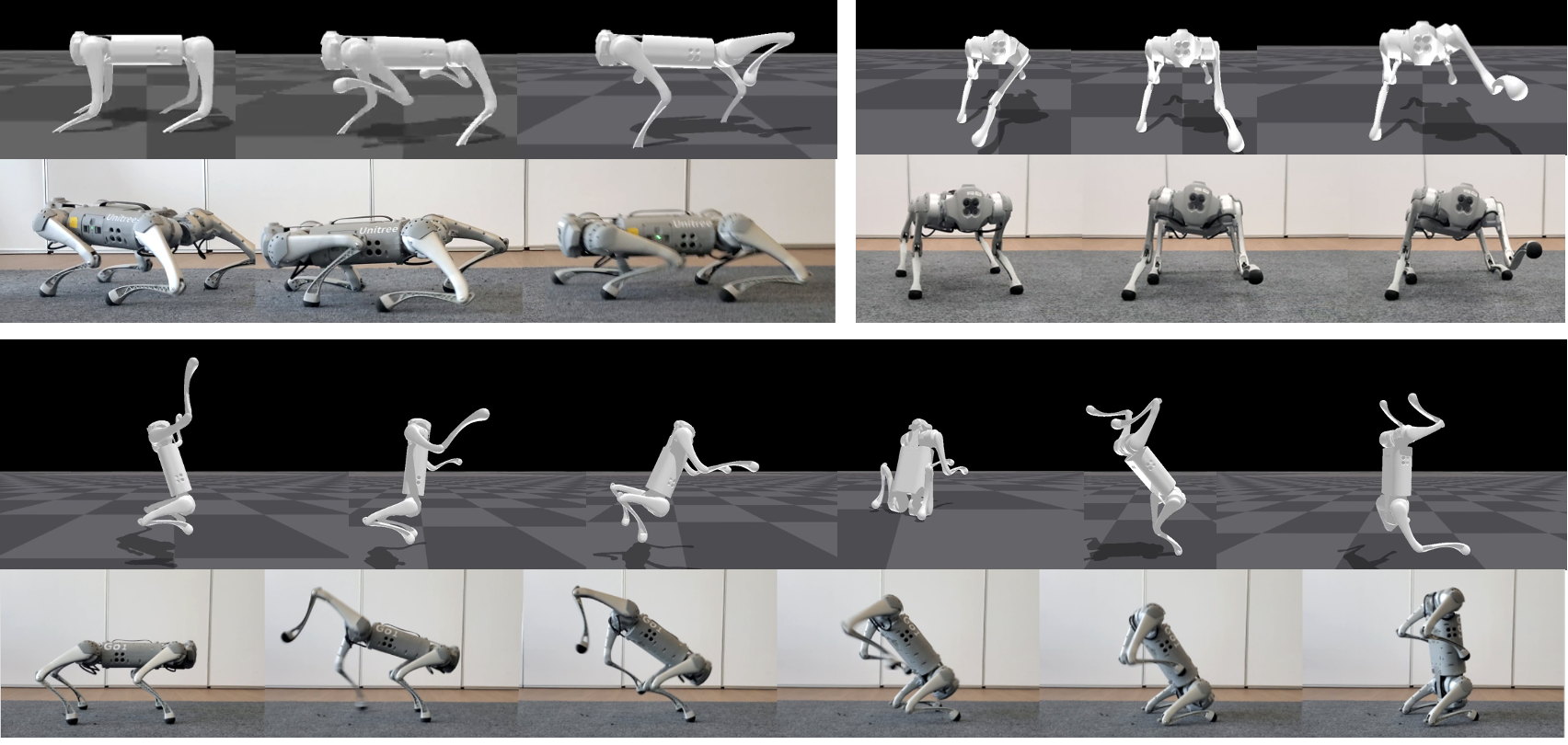}
    \vspace{-5mm}
    \caption{The reference motions and the behaviors of the real-world robot. The reference motions are physics-ignorant.
    }
    \label{fig:real}
    \vspace{-4mm}
\end{figure}

We demonstrate the real-world results of four different motions in Fig.(\ref{fig:teaser},~\ref{fig:real}), including walking backward, raising the left hand, handstand, and throwing a ball. 

We have identified multiple options for retargeting strategies, and {\name} works well for \emph{all} of these strategies and generates diverse control policies. For instance, in the case of the commands "handstand" and "throw a ball," we retarget all the robot joints, aligning the human hands with the robot's front legs and the human legs with the robot's rear legs (refer to Fig.~\ref{fig:joint_mapping}). Consequently, the robot executes these actions like a human. For the "walk backward" command, we realign the human legs with the robot's rear legs and mirror the states to the front legs, allowing the robot to mimic a dog's walking pattern. Furthermore, we also explored retargeting of partial joints. When given the command "raise the left hand," we solely retarget the joints of the left front hand. As a result, the robot stands on three legs while extending its left hand.

We present a comparison between the retargeted reference motions and the actual behaviors of the real-world robot in Fig.~\ref{fig:real}. Even though the reference motions appear to defy the laws of physics, our {\name} policies excel at completing these tasks.
For example, consider the command "walk backward." In the reference motion, the posture depicted cannot maintain stability, but our robot executes a secure backward walk. Similarly, when instructed to perform a "handstand," the reference motion begins from a standing posture and concludes in a precarious state. In contrast, our policy achieves a safe and steady handstand.

%% file: table.tex
\begin{table}[]
    \centering
    \scriptsize
    \setlength{\tabcolsep}{4pt}
    \vspace{1mm}
    \caption{Success rates of different tasks. We train a single policy over all 4 terrains and evaluate the policy separately on each terrain. We train the policies over 3 seeds. }
    \vspace{-2mm}
    \begin{tabular}{c|cccc}
    \toprule
    Terrian & Plane & Rand. & Pyramid & Wave\\
    \midrule
    \multicolumn{5}{c}{Task: Cartwheel}\\
    \midrule
    {\name} & $\mathbf{100.0 \pm 0.0}$ & $\mathbf{98.8 \pm 0.5}$ & $\mathbf{85.2 \pm 6.6}$ & $\mathbf{86.4 \pm 10.6}$ \\
    GAILfO & $66.7 \pm 47.1$ & $65.6 \pm 46.3$ & $33.1 \pm 38.7$ & $58.7 \pm 34.7$ \\
    RGAILfO & $0.0 \pm 0.0$ & $6.3 \pm 8.6$ & $16.3 \pm 22.7$ & $19.9 \pm 13.2$ \\
    State Err. & $0.0 \pm 0.0$ & $4.7 \pm 6.6$ & $15.0 \pm 21.1$ & $11.9 \pm 15.3$ \\
    \midrule
    Pure RL & - & - & - & - \\
    BCO & $0.0 \pm 0.0$ & $0.0 \pm 0.0$ & $0.0 \pm 0.0$ & $0.0 \pm 0.0$ \\

    \midrule
    \multicolumn{5}{c}{Task: Kick}\\
    \midrule
    {\name} & $\mathbf{100.0 \pm 0.0}$ & $\mathbf{78.9 \pm 6.8}$ & $\mathbf{89.9 \pm 7.0}$ & $69.7 \pm 7.5$ \\
    GAILfO & $0.0 \pm 0.0$ & $46.6 \pm 17.5$ & $60.4 \pm 15.1$ & $45.5 \pm 8.3$ \\
    RGAILfO & $33.3 \pm 47.1$ & $28.9 \pm 18.9$ & $39.7 \pm 14.2$ & $29.5 \pm 14.0$ \\
    State Err. & $\mathbf{100.0 \pm 0.0}$ & $75.3 \pm 24.9$ & $85.4 \pm 20.7$ & $\mathbf{70.0 \pm 29.2}$ \\
    \midrule
    Pure RL & $100.0 \pm 0.0$ & $100.0 \pm 0.0$ & $100.0 \pm 0.0$ & $100.0 \pm 0.0$ \\
    BCO & $0.0 \pm 0.0$ & $0.0 \pm 0$ &  $0.0 \pm 0.0$ & $0.0 \pm 0.0$ \\

    \midrule
    \multicolumn{5}{c}{Task: Run Backwards}\\
    \midrule
    {\name} & $\mathbf{100.0 \pm 0.0}$ & $99.8 \pm 0.2$ & $\mathbf{100.0 \pm 0.0}$ & $95.5 \pm 1.2$ \\
    GAILfO & $\mathbf{100.0 \pm 0}$ & $99.6 \pm 0.4$ & $99.9 \pm 0.1$ & $96.2 \pm 1.7$ \\
    RGAILfO & $\mathbf{100.0 \pm 0.0}$ &  $\mathbf{99.9 \pm 0.0}$ & $99.8 \pm 0.3$ & $\bf{98.8 \pm 0.5}$ \\
    State Err. & $\mathbf{100.0 \pm 0.0}$ &  $99.1 \pm 0.4$ & $\mathbf{100.0 \pm 0.0}$ & $97.5 \pm 1.0$  \\
    \midrule
    Pure RL & $97.2 \pm 0.7$ & $25.5 \pm 2.1$ & $96.6 \pm 0.2$ & $24.0 \pm 0.2$  \\
    BCO & $14.0 \pm 0.5$ & $13.6 \pm 0.4$ & $12.8 \pm 0.9$ & $13.5 \pm 0.9$ \\

    \bottomrule
    \end{tabular}
    \label{tab.compare}
    \vspace{-6mm}
\end{table}

%% file: 60_con.tex
\section{Conclusion}
We propose a multi-phase method {\name} to train robot control policy following the given language command. We first generate human motion using a language-conditioned motion diffusion model, and retarget the generated human motion to the robot skeleton. We adopt RL to train control policies. {\name} finally produces a robust policy that controls a real-world quadrupedal robot to take behaviors consistent with the language commands. 


%% file: appendix.tex
\appendix

\section{Implementation Details}
\label{appendix.details}

\subsection{Motion Generation Details}
\label{appendix.mdm}
We use the official implementation\footnote{\url{https://github.com/GuyTevet/motion-diffusion-model/}} of Human Motion Generation Model~\cite{tevet2022human} to generate motion sequences. 
They use the CLIP model to encode natural language descriptions and use a transformer denoiser over a 263-dimensional feature space proposed in~\cite{guo2022generating}. The generated feature is then rendered into SMPL mesh~\cite{loper2015smpl}. In the sampling process, following their default hyper-parameters, we adopt the classifier-free guidance~\cite{ho2022classifier} with a guidance scale of $2.5$. We use cosine noise scheduling with 1000 diffusion steps.

\begin{figure*}[ht]
    \centering
    \begin{minipage}[t]{.48\textwidth}
        \centering
        \includegraphics[width=\textwidth]{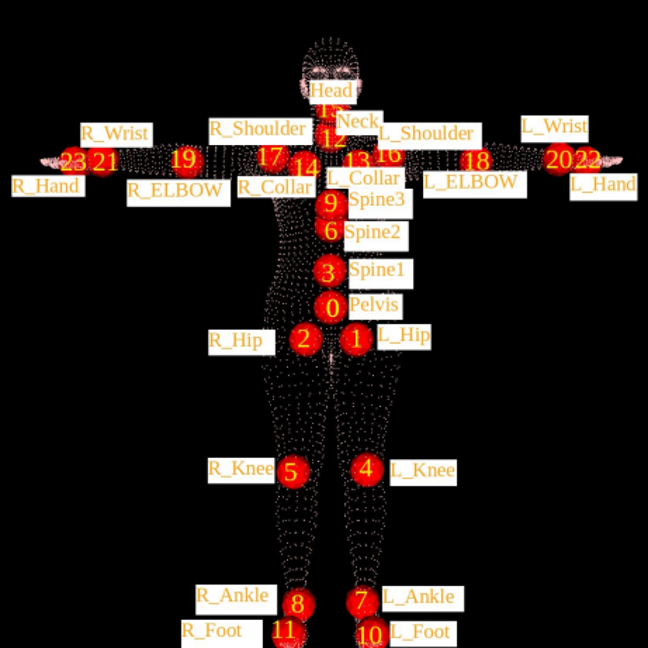}
        \caption{Skeleton layout of the SMPL~\cite{loper2015smpl} model. This image comes from the \textit{SMPL make simple} tutorial (\url{https://smpl-made-simple.is.tue.mpg.de/}).}
        \label{fig:smpl}
    \end{minipage}
    \begin{minipage}[t]{.48\textwidth}
        \centering
        \includegraphics[width=\textwidth]{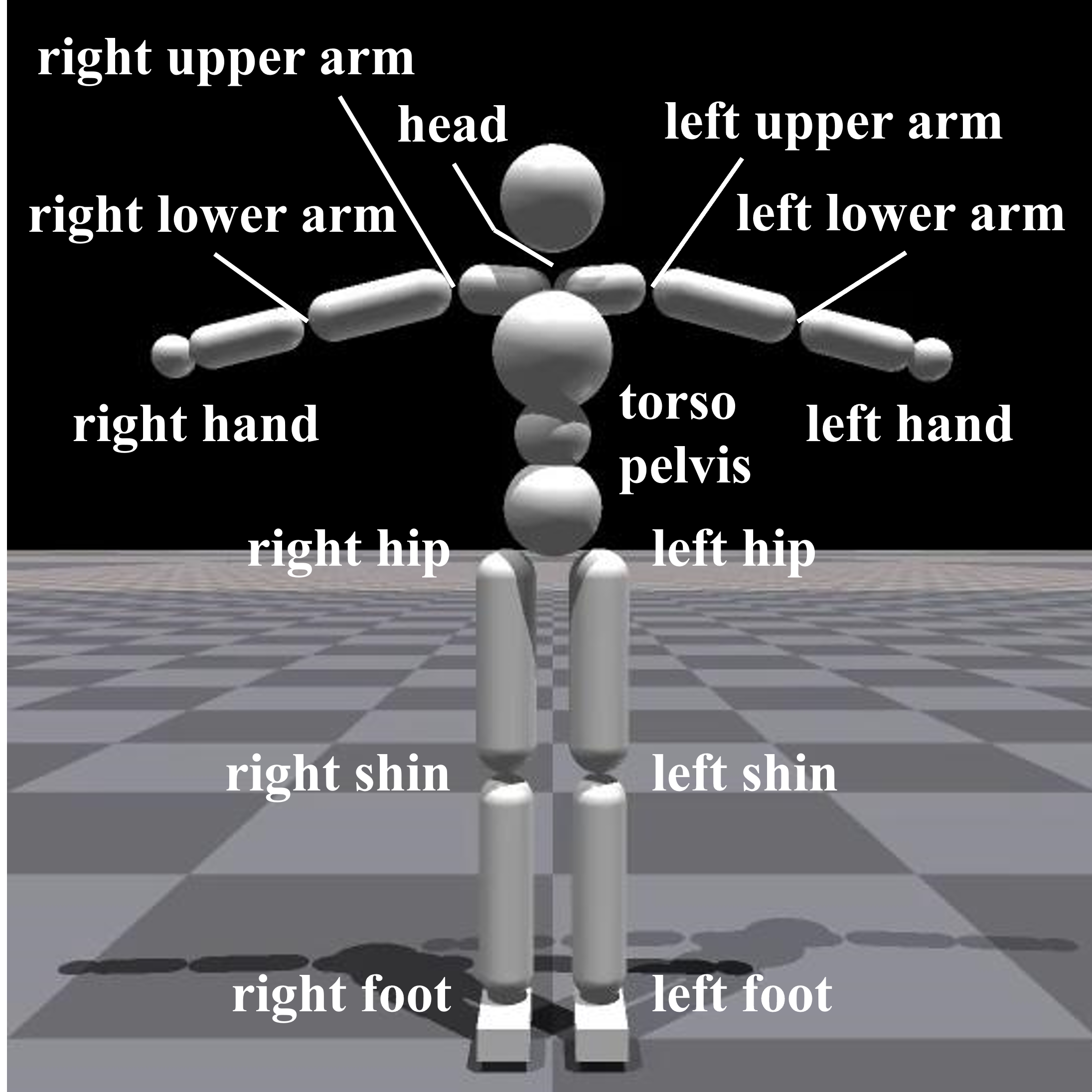}
        \caption{Skeleton layout of the humanoid robot.}
        \label{fig:humanoid_robot}
    \end{minipage}
\end{figure*}

\subsection{Retargeting Details}
\label{appendix.retraget}
We retarget the generated SMPL meshes to our target skeletons (humanoid or quadrupedal robot) using a script\footnote{\url{https://github.com/NVIDIA-Omniverse/IsaacGymEnvs/tree/main/isaacgymenvs/tasks/amp/poselib}} from Isaac Gym~\cite{makoviychuk2021isaac,peng2021amp}. First, we extract the root translation and local joint rotations from SMPL parameters. As presented in Fig.~\ref{fig:smpl} and Fig.~\ref{fig:humanoid_robot}, the SMPL skeleton layout has more joints than our humanoid robot. We remove redundant joints and map the rest according to the mapping given in Tab.~\ref{tab:joint_mapping} and Tab.~\ref{tab:joint_mapping_dog}. Then we compute the relative joint rotations compared to a T-pose of the source motion. These rotations are applied on the target skeleton to obtain the retargeted motion.

For the humanoid robot, the elbows and knees (labeled ``lower arm'' and ``shin'' in Fig.~\ref{fig:humanoid_robot}) have only 1 DoF instead of 3 in the SMPL skeleton. We convert these 3D joints to 1D using inverse dynamics. Take the left arm as an example. We first compute the relative rotation angle between the left upper arm and the left lower arm and the local rotation of the left elbow is set to this angle. The local rotation of the left shoulder is then adjusted such that the position of the left hand stays in the original position.

\begin{figure*}[ht]
    \centering
    \includegraphics[width=.98\textwidth]{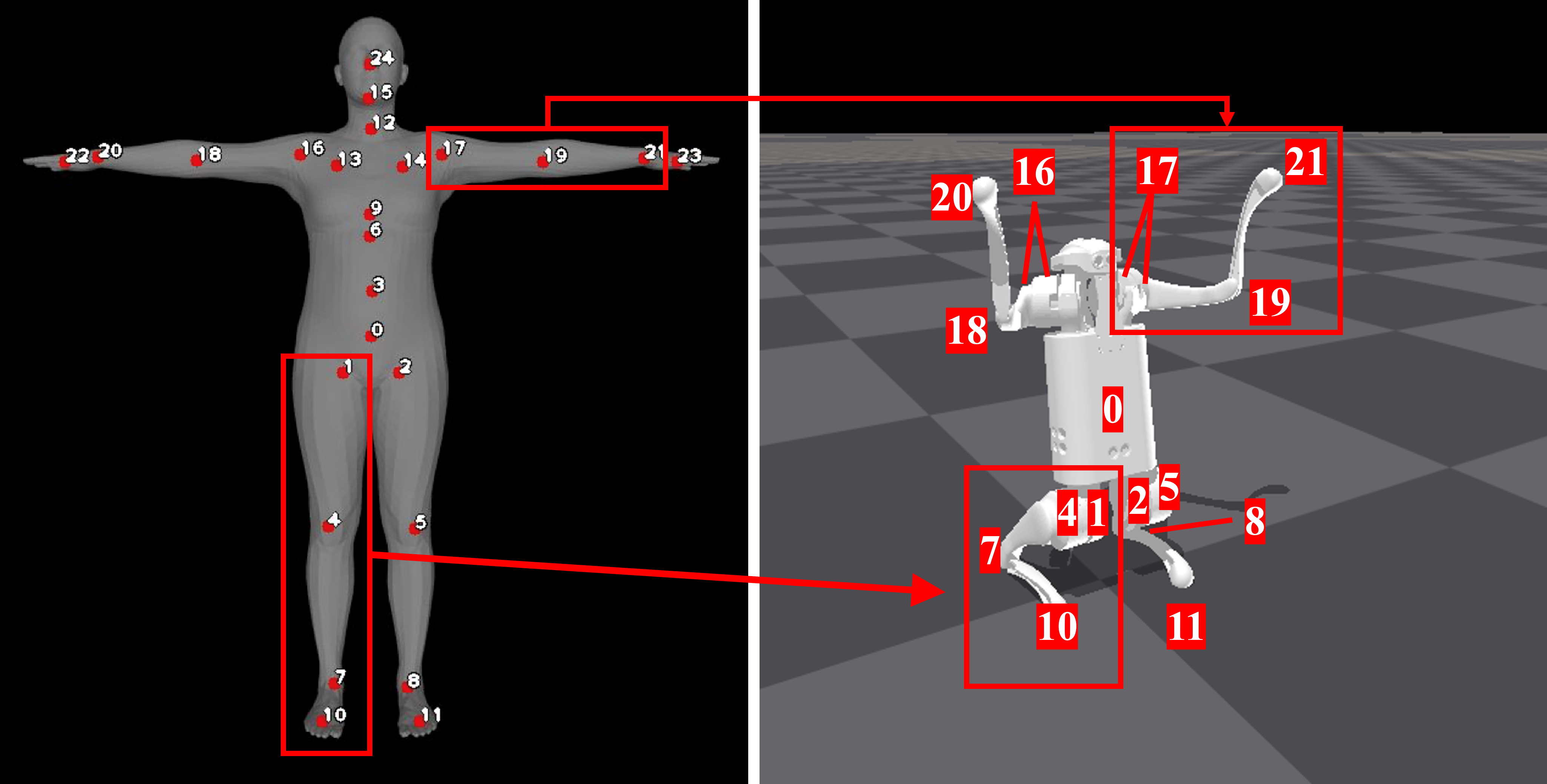}
    \caption{Joint mappings from the SMPL skeleton to the quadrupedal robot. Note that 16 and 17 in the quadrupedal skeleton each correspond to 2 joints. The joint names are listed in Tab.~\ref{tab:joint_mapping_dog}}
    \label{fig:joint_mapping_dog}
\end{figure*}

The four legs of the quadrupedal robot have a similar structure. Take the right rear leg as an example. The hip, thigh, and calf (labeled 1, 4, and 7 in Fig.~\ref{fig:joint_mapping_dog}) have 1 DoF each. We apply different rules to retarget the arms and legs of an SMPL character to the front and rear legs of a quadrupedal robot.
For the rear legs, the hips, knees, and ankles of the SMPL character are mapped to the hips, thighs, and calves of the quadrupedal robot, respectively. Redundant DoFs are discarded in this process. For the front legs, the elbow and wrist of the SMPL character are mapped to the calf and foot of the quadrupedal robot. The shoulder of the SMPL character has 3 DoFs. We map one of them to the front hip and one to the front thigh of the quadrupedal robot. The joint mappings are visualized in Fig.~\ref{fig:joint_mapping_dog}.

\begin{table}[ht]
    \centering
    \begin{tabular}{|l|l|}
\hline
Source Joints & Target Joints \\
\hline
Pelvis & pelvis \\
L\_Hip & left\_thigh \\
L\_Knee & left\_shin \\
L\_Ankle & left\_foot \\
R\_Hip & right\_thigh \\
R\_Knee & right\_shin \\
R\_Ankle & right\_foot \\
Spine3 & torso \\
Neck & head \\
L\_Shoulder & left\_upper\_arm \\
L\_Elbow & left\_lower\_arm \\
L\_Wrist & left\_hand \\
R\_Shoulder & right\_upper\_arm \\
R\_Elbow & right\_lower\_arm \\
R\_Wrist & right\_hand \\
\hline
    \end{tabular}
    \caption{Joint mapping between the SMPL skeleton (Source Joints) and the humanoid robot (Target Joints).}
    \label{tab:joint_mapping}
\end{table}

\begin{table}[ht]
    \centering
    \begin{tabular}{|l|l|c|}
\hline
Source Joints & Target Joints & Joint Index \\
\hline
Pelvis & trunk & 0 \\
L\_Hip & RL\_hip & 2 \\
L\_Knee & RL\_thigh & 5 \\
L\_Ankle & RL\_calf & 8 \\
L\_Foot & RL\_foot & 11 \\
R\_Hip & RR\_hip & 1 \\
R\_Knee & RR\_thigh & 4 \\
R\_Ankle & RR\_calf & 7 \\
R\_Foot & RR\_foot & 10 \\
L\_Shoulder & FL\_hip, FL\_thigh & 17 \\
L\_Elbow & FL\_calf & 19 \\
L\_Wrist & FL\_foot & 21 \\
R\_Shoulder & FR\_hip, FR\_thigh & 16 \\
R\_Elbow & FR\_calf & 18 \\
R\_Wrist & FR\_foot & 20 \\
\hline
    \end{tabular}
    \caption{Joint mapping between the SMPL skeleton (Source Joints) and the quadrupedal robot (Target Joints). ``Target Joint Index'' indicates the corresponding indicies in Fig.~\ref{fig:joint_mapping_dog}.}
    \label{tab:joint_mapping_dog}
\end{table}

\subsection{Training Details}
\label{appendix.rl_details}

\subsubsection{Reinforcement Learning}
We adopt Isaay Gym~\cite{makoviychuk2021isaac} as our physics simulator, as it enables massive parallel simulation on GPUs. For the humanoid robot, the simulation runs at 60Hz while the frequency of the control policy is 30Hz. The control policy $\pi_\theta$, the value function $V_\phi$, and the discriminator $D_\psi$ are MLPs with hidden dimensions (1024, 512) and ReLU activations.
We adopt a fixed diagonal covariance matrix $\Sigma$ for the control policy,  with all diagonal entries set to 0.05. We normalize the policy’s input state using a running estimate of the mean and variance of the states. We create 4096 parallel simulation environments in Isaac Gym to collect training samples. The max episode length of each simulation is 300. We update $\pi_\theta$, $V_\phi$, and $D_\psi$ 6 times every 16 steps of the environments, with a mini-batch size of 32768. The clipping coefficient of PPO is set to 0.2. The discount $\gamma$ of the MDP is set to 0.99. We also adopt GAE~\cite{schulman2015high} to estimate the advantage of policy gradient and the GAE coefficient is set to 0.95. We adopt Adam~\cite{kingma2014adam} as the optimizer and set the learning rate as 5e-5. We train the policy for 5000 epochs. All the experiments on humanoid robots use the above hyper-parameters. For \name{}, we set both $\lambda_{adv}$ and $\lambda_{err}$ to 1, except for the task ``run backward'' where $\lambda_{err}$ is set to $0.1$.

For the experiments on the quadrupedal robot, the simulation runs at 200Hz and the control policy runs at 50Hz. We run the experiments with 8192 environments for 20,000 epochs. The maximum episode length is set to 250. Other hyper-parameters are the same as the experiments on the humanoid robot.

We employ the technique of early termination~\cite{peng2018deepmimic,peng2021amp} to accelerate training. For the humanoid robot, we reset the environment when any joint of the humanoid, with the exception of feet, has a nonzero contact force. For the task ``cartwheel'', the hands of the humanoid are also excluded. For the quadrupedal robot, an environment is reset if any joint rotation or angular velocity exceeds a certain limit. It is possible that the quadrupedal robot falls on the ground without triggering an early termination. We make sure that a quadrupedal robot in such a state receives a low reward on state similarity. The details about reward design will be described in Appendix.~\ref{appendix.state_similarity}.

\subsubsection{States}
For the humanoid robot, the state consists of:
\begin{itemize}
    \item The root rotation, linear velocity, and angular velocity.
    \item Local rotations and angular velocities of each joint.
    \item Positions of the hands and feet.
\end{itemize}

For the quadrupedal robot, the root’s linear velocity and angular velocity could not be estimated precisely by the real-world go1 robot. Following~\cite{rudin2022learning}, we utilize the gravity projected on the robot’s up axis and actions taken in the last timestep. The state consists of:
\begin{itemize}
    \item Projected gravity.
    \item Local rotations and angular velocities of each joint.
    \item Positions of the hands and feet.
    \item Actions taken in the last timestep.
\end{itemize}

\subsubsection{State Similarity}
\label{appendix.state_similarity}
For the humanoid robot, we use the reward function in \cite{peng2018deepmimic} as our state similarity metric:
\begin{equation}
\begin{split}    
\label{eq.err_rew}
    Sim(y,s) &= 0.65 r_p + 0.1 r_v + 0.15 r_e + 0.1 r_r \\
    r_p &= \exp\left[ -2 \sum_j(\|q_{s,j} - q_{y,j} \|^2) \right] \\
    r_v &= \exp \left[ -0.1 \sum_j(\|\dot{q}_{s,j} - \dot{q}_{y,j} \|^2) \right] \\
    r_e &= \exp \left[ -40 \sum_e(\| p_{s,e} - p_{y,e} \|^2) \right] \\
    r_r &= \exp \left[ -10 \| p_{s,r} - p_{y,r} \|^2 \right] 
\end{split}
\end{equation}
Here $q_{\cdot,j}$ and $\dot q_{\cdot,j}$ denote the local rotation and angular velocity of the $j$-th joint. $p_{\cdot,j}$ denotes the relative position to root of the $j$-th joint. We use relative positions to root instead of absolute positions since the humanoid robot may be initialized at random positions. $p_{\cdot,r}$ denotes the global root position.

For the quadrupedal robot, we additionally take into account the root rotation. Let $q_{\cdot,r}$ denote the root rotation. The similarity metric is changed to
\begin{multline}
    Sim(y,s)=0.8r_\mathrm{rot}(0.65 r_p + 0.1 r_v + 0.15 r_e + 0.1 r_r) \\
    +0.2r_\mathrm{rot},
\end{multline}
where $r_\mathrm{rot}=\max(0,q_{s,r}\cdot q_{y,r})$. The $r_\mathrm{rot}$ term penalizes incorrect root rotations. Thus, when the quadrupedal robot falls over, it receives a low reward even if local rotations of other joints still match the reference motion.

\subsubsection{State Matching}
As dynamic programming can be time-consuming, we maintain a state matching $(u_{0:k},v_{0:k})$ and update it every certain number of epochs. The state matching is initialized as the simplest matching $u_i=v_i=i$. For the task ``run back'' and ``kick'', the matching is updated every 4000 epochs. For the task ``cartwheel'', the matching is updated every 1000 epochs. During each update, we collect 4096 episodes and use dynamic programming to compute a matching with the highest total reward for each episode. We filter out matched pairs of states with similarity less than $0.05$. Then we select the episode with the most states matched. As we will initialize the humanoid robot from two different states, we compute a matching for each state initialization. 

Another effective matching for the task ``run back'' and ``kick'' is to simply match the initial states of the reference motion. More specifically, for a reference motion $y_{0:H}$ and a trajectory $s_{0:T}$ from the policy, we simply set $u_i=i$ and $v_i=i+t_0$, where $t_0=\min\{t\mid Sim(y_0,s_t)\ge 0.5\}$. This matching strategy is also effective for experiments on the quadrupedal robot.


\subsection{Domain Randomization}
\label{appendix.domain}
For the humanoid robot, we use the default domain randomization parameters in Isaac Gym. We list the randomized physics parameters in Tab.~\ref{tab:humanoid_rand}. Except for observation, action, and rigid body mass, all the parameters are linearly interpolated between no randomization and max randomization in the first 3000 environment steps. Due to a current limitation of Isaac Gym\footnote{\url{https://github.com/NVIDIA-Omniverse/IsaacGymEnvs/blob/main/docs/domain_randomization.md}}, rigid body mass is randomized only once before the simulation is started. New randomizations are generated every 600 environment steps. We also randomize the terrain when training the humanoid robot. There are four terrains during training. “Plane” refers to a flat surface without variations in elevation.
For the ``Rand'' terrain, we generate a height field where the height of a vertex is sampled from $\{0, 0.2\text{m}\}$. The horizontal distance between two adjacent points is 0.5m. This height field is then converted into a triangle mesh. The ``Pyramid'' terrain is a square cone with steps. The step width is 0.5m and the step height is 0.05m. There is a flat platform at the peak of the pyramid with a side length of 1m. The ``Wave'' terrain is constructed from sine waves. More specifically, the height at $(x,y)$ is given by $0.5\cos(\pi x/6)+0.5\sin(\pi y/6)$ (in meters). Each terrain has a size of $240\text{m}\times 240\text{m}$. The humanoid robots are randomly initialized in the center $120\text{m}\times 120\text{m}$ region. This ensures that the robots do not leave the terrain within an episode.

\begin{table}[ht]
    \centering
    \begin{tabular}{|c|c|c|c|c|}
        \hline
        Parameter & Operation & Distribution & Unit \\
        \hline
        Observation & Additive & $\mathcal N(0,0.002)$ & - \\
        Action & Additive & $\mathcal N(0,0.02)$ & - \\
        Gravity & Additive & $\mathcal N(0,0.4)$ & $\mathrm m/\mathrm s^2$ \\
        Body Mass & Scaling & $\operatorname{U}(0.5,1.5)$ & 1 \\
        Body Friction & Scaling & $\operatorname{U}(0.7,1.3)$ & 1 \\
        Body Restitution & Scaling & $\operatorname{U}(0,0.7)$ & 1 \\
        Damping & Scaling & $\operatorname{U}(0.5,1.5)$ & 1 \\
        Stiffness & Scaling & $\operatorname{U}(0.5,1.5)$ & 1 \\
        Lower & Additive & $\mathcal N(0,0.01)$ & $\mathrm{rad}$ \\
        Upper & Additive & $\mathcal N(0,0.01)$ & $\mathrm{rad}$ \\
        \hline
    \end{tabular}
    \caption{Randomization parameters of the humanoid robot. $U(a,b)$ denotes the uniform distribution over $[a,b]$. ``Lower'' and ``Upper'' refer to the lower bounds and upper bounds of the range of the DoFs.}
    \label{tab:humanoid_rand}
\end{table}

For the quadrupedal robot, because of the limitation of Isaac Gym, we fix the ground friction coefficient and randomize the friction coefficients of the robot's bodies. We train the policy on the plane terrain and set the friction and restitution coefficients of the ground as 0.6 and 0.4, respectively. We randomize the mass of the robot trunk and the friction of all the rigid bodies. We also randomize the proportional gain and derivative gain of the PD controller. In addition, we noise the observations. The detailed ranges are listed in Tab~\ref{tab:dog_rand}.

\begin{table}[ht]
    \centering
    \small
    \begin{tabular}{|c|c|c|c|c|}
        \hline
        Parameter & Operation & Distribution & Unit \\
        \hline
        Obs.Gravity & Additive & $\operatorname{U}(-0.05,0.05)$ & $\mathrm m/\mathrm s^2$ \\
        Obs.ROT & Additive & $\operatorname{U}(-0.01, 0.01)$ & $\mathrm{rad}$\\
        Obs.VEL & Additive & $\operatorname{U}(-1.5, 1.5)$ & $\mathrm{rad}/\mathrm s$ \\
        Trunk Mass & Additive & $\operatorname{U}(0, 1)$ & $\mathrm{kg}$ \\
        Body Friction & Scaling
        & $\operatorname{U}(0.05, 2)$ & 1 \\
        Proportional Gain & Scaling & $\operatorname{U}(0.7, 1.3)$ & 1 \\
        Derivative Gain & Scaling & $\operatorname{U}(0.7,1.3)$ & 1 \\
        \hline
    \end{tabular}
    \caption{Randomization parameters of the quadrupedal robot. $U(a,b)$ denotes the uniform distribution over $[a,b]$. We noise the policy inputs separately. Obs.Gravity refers to the projected gravity, Obs.ROT and Obs.VEL refers to the local rotations and velocities.}
    \label{tab:dog_rand}
\end{table}

\section{Other Results on Quadrupedal Robot}
In this section, we demonstrate more experiments on the quadrupedal robot. Please also refer to our videos.
\subsection{Partial Retargeting}
We conduct experiments on the quadrupedal robot by retargeting a subset of joints.

\begin{figure}[ht]
    \centering
    \includegraphics[width=\linewidth]{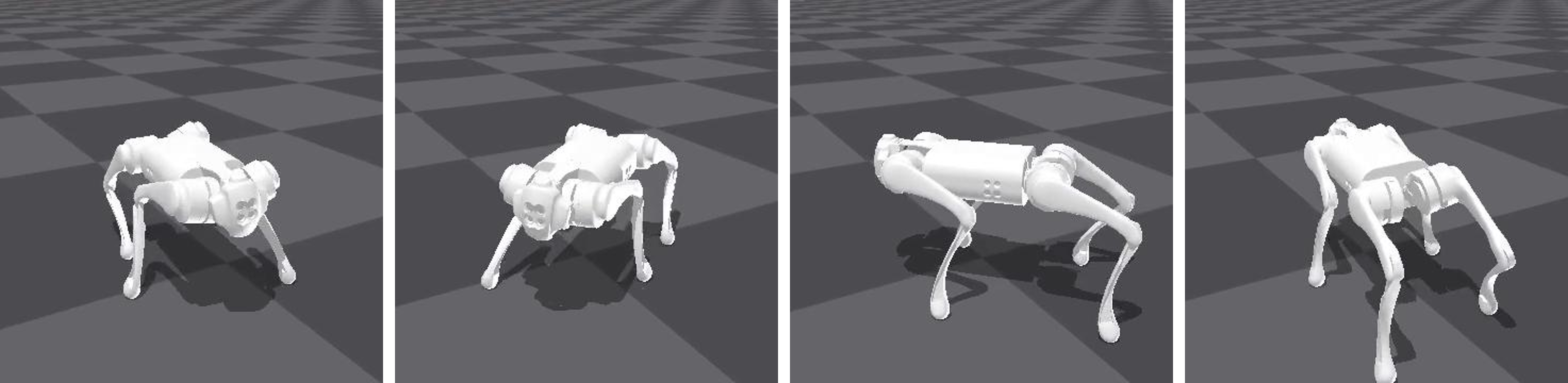}
    \caption{The quadrupedal robot is turning around. Note that only the root rotation is retargeted.}
    \label{fig:quad_turn_around}
\end{figure}

\paragraph{Turn around}
We generate a motion sequence using the language prompt ``turn around'' and only retarget the root rotation. As presented in Fig.~\ref{fig:quad_turn_around}, the quadrupedal robot learns to turn around in the physics simulator. 

\begin{figure}[ht]
    \centering
    \includegraphics[width=0.5\linewidth]{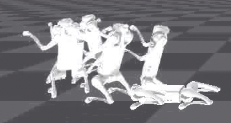}
    \caption{``Turn around''. We also retarget all the bodies when mimicking the reference motion of ``turning around.'' which makes the quadruped robot walk on its two rear
    legs and turn around.}
    \label{fig:turn_v2}
\end{figure}

It is important to
note that we intended to show the flexibility and versatility of our approach by employing various retargeting strategies. We can also retarget all the bodies when mimicking the reference motion of “turning around.”(Fig.\ref{fig:turn_v2}), which makes the quadruped robot walk on its two rear
legs and turn around. 

\begin{figure}[ht]
    \centering
    \includegraphics[width=.8\linewidth]{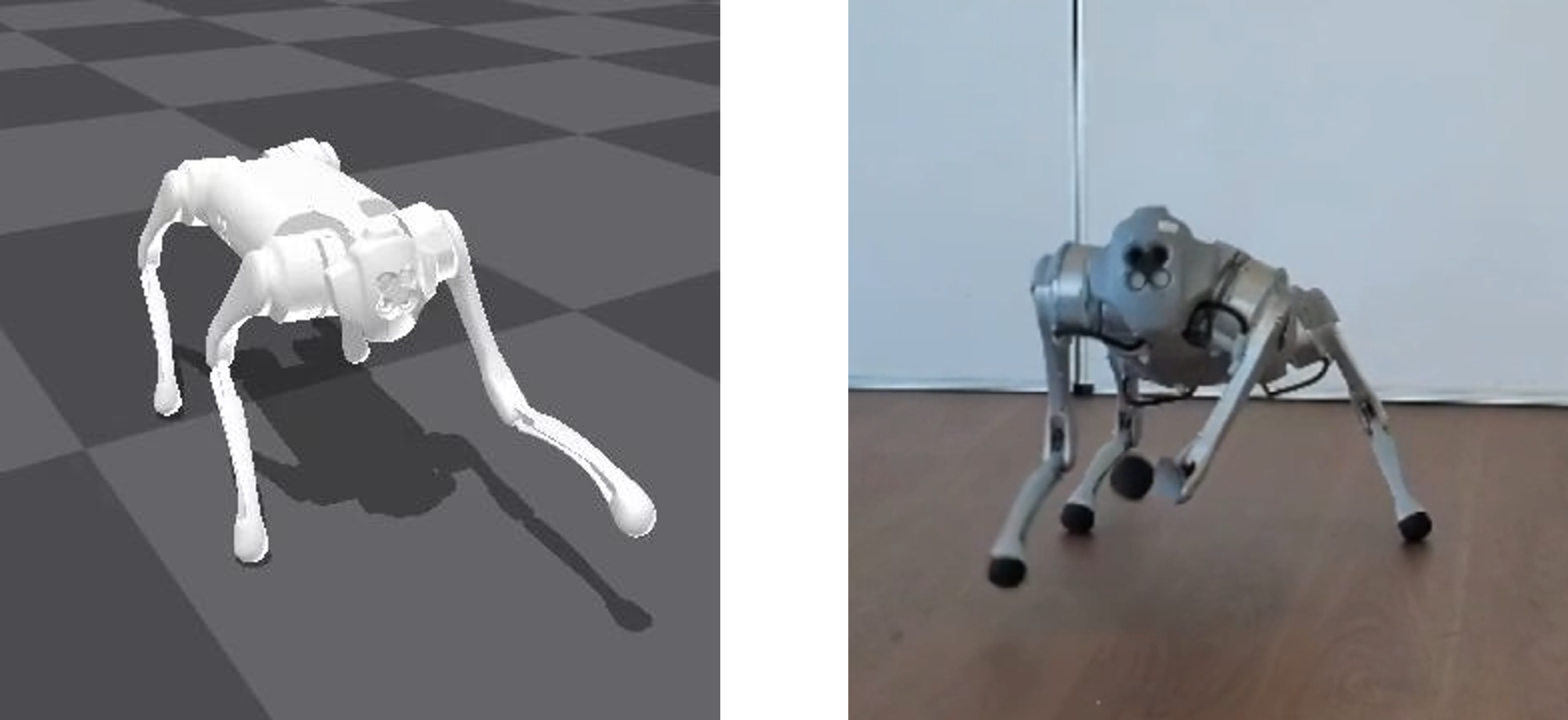}
    \caption{(Left) We only retarget the left arm of the SMPL character to the left front leg of the quadrupedal robot. (Right) The policy learns to raise the left front leg in the real world.}
    \label{fig:quad_raise_hand}
\end{figure}

\paragraph{Raise the left hand}
We generate another motion sequence of a person raising the left hand by the language prompt ``raise the left hand''. We retarget only the left arm to the left front leg of the quadrupedal robot. As shown in Fig.~\ref{fig:quad_raise_hand}, the robot raises its left front leg with the other three legs on the ground in the reference motion. The learned policy can perform a similar behavior in the real world.

